\documentclass[journal]{IEEEtran}
\usepackage{verbatim}
\usepackage{booktabs} 
\usepackage{amsmath}
\usepackage{multicol, blindtext}
\usepackage{algorithm}
\usepackage[noend]{algpseudocode}
\usepackage{courier}
\usepackage{amsmath}
\algdef{SE}[DOWHILE]{Do}{DoWhile}{\algorithmicdo}[1]{\algorithmicwhile\ #1} 
\usepackage{graphicx}
\makeatletter
\def\BState{\State\hskip-\ALG@thistlm}
\makeatother
\usepackage{caption,subcaption}
\usepackage{textcomp}
\usepackage{multirow}
\usepackage[latin1]{inputenc}
\usepackage{tikz}
\usepackage{makecell}
\usepackage{colortbl}
\xdefinecolor{gray95}{gray}{0.65}
\xdefinecolor{gray25}{gray}{0.8}
\usetikzlibrary{shapes,arrows}
\tikzstyle{decision} = [diamond, draw, fill=white!20, 
    text width=4.5em, text badly centered, node distance=3cm, inner sep=0pt]
\tikzstyle{block_init} = [rectangle, draw, fill=white!20, 
    text width=9em, text centered, rounded corners, minimum height=2em]
\tikzstyle{block} = [rectangle, draw, fill=white!20, 
    text width=5em, text centered, rounded corners, minimum height=2em]
\tikzstyle{cblock} = [rectangle, draw, fill=gray!20, 
    text width=5em, text centered, rounded corners, minimum height=2em]
\tikzstyle{line} = [draw, -latex']
\tikzstyle{cloud} = [draw, ellipse,fill=white!20, node distance=3cm,
    minimum height=2em]
\usepackage{mathtools}

\ifCLASSINFOpdf
\else
\fi
\begin{document}
%
\title{Performance Analysis and Improvement of Parallel Differential Evolution}
%
%
%

\author{Zibin Pan
	\\ The Chinese University of Hongkong, Shenzhen
	\\ zibinpan@link.cuhk.edu.cn

}

\newcommand{\eg}[1]{{\color{green}{#1}}}

\maketitle

\begin{abstract}
Differential evolution (DE) is an effective global evolutionary optimization algorithm using to solve global optimization problems mainly in a continuous domain. In this field, researchers pay more attention to improving the capability of DE to find better global solutions, however, the computational performance of DE is also a very interesting aspect especially when the problem scale is quite large. Firstly, this paper analyzes the design of parallel computation of DE which can easily be executed in Math Kernel Library (MKL) and Compute Unified Device Architecture (CUDA). Then the essence of the exponential crossover operator is described and we point out that it cannot be used for better parallel computation. Later, we propose a new exponential crossover operator (NEC) that can be executed parallelly with MKL/CUDA. Next, the extended experiments show that the new crossover operator can speed up DE greatly. In the end, we test the new parallel DE structure, illustrating that the former is much faster.
\end{abstract}


\IEEEpeerreviewmaketitle

\section{Introduction}

\IEEEPARstart{D}{ifferential} evolution (DE) is a simple and useful evolutionary algorithm which is similar to the Genetic Algorithm (GA) and able to address optimization problems more efficiently and precisely \cite{Opara2019, Pham2011}. Since DE was proposed by Price and Storn in 1995 with several papers \cite{Storn1995, Price1996, Storn1996}, it has attracted a lot of people to use and do further researches because of its simple realization, wide application, and good optimization results.

Most researchers pay more attention to reforming DE in order to make it solve difficult problems better. But as for some large-scale optimization problems, although the evaluation of individuals is the bottleneck of the computational performance most of the time, we still found that if we use CPU/GPU parallel technologies in DE, we can get a dramatic increase in the performance.

In DE, each chromosome is an array containing a list of real numbers. This simple and useful structure makes it possible to calculate parallelly by matrix calculation techniques with the help of Math Kernel Library(MKL) or Compute Unified Device Architecture (CUDA). The former one performs well in matrix calculation with CPU while the other one can speed up the code with Graphics Processing Unit (GPU).

Until now there are many achievements in this field, for example, DE with GPGPU (General Purpose GPU), is a parallel version of DE executing in GPU, introduced by de Veronese and Krohling \cite{Veronese2010}, Zhu \cite{Zhu2011}, and Zhu and Li \cite{Zhu2010}. Cortes et.al \cite{Cortes2015} analyzed the influence of parameters on speedup and the quality of solutions of DE on a GPGPU, showing that DE on a GPGPU is not only running faster but also has a good optimization result. Recently, Meselhi et. al. designed a fast differential evolution for big optimization \cite{Meselhi2018}, which shows a big increase in the computational performance.

However, these parallel calculation achievements can only perform well in "DE/*/*/bin". They don't do well in some other versions of DE such as "DE/*/*/L", which uses the exponential crossover (exp) instead of the uniform (binomial) crossover in recombination. The work of Tanabe \cite{Tanabe2014} showed that the exp crossover can perform better in some optimization problems where the adjacent variables have some dependencies. Hence, it is still useful in DE. However, in this context, it will show that the exp crossover cannot be calculated parallelly enough as the uniform (binomial) crossover so that when using parallel calculation the former one runs much slower than the latter.

The rest of the paper is organized as follows. Section 2 shows several parallel-computational algorithms in a different part of DE. In section 3, it will uncover the essence of exponential crossover and changed it with the help of the probability theory so that it can be faster as well as be able to speed up on MKL/GPU. In section 4, we will compare the performances of DE, DE on MKL (MKL-DE), and DE with CUDA (CUDA-DE).

\section{Parallel Differential Algorithm}

The Differential evolutionary algorithm processes the population with $N$ individuals. Each individual owns a chromosome which is an $n$-dimension vector $x_{i,g}$ made of real numbers. $i$ is the index number of the individual in the population, $g$ indicates the generation to which a vector belongs. $x_g$ is a matrix that stores all the $x_{i,g}$ in the population. After initialization, base vector selection, differential mutation, recombination, and selection, a new generation is created and after several iterations, the solutions of the population can get better and better. The pseudocode of DE is shown in Algorithm \ref{alg:DE}.
\begin{algorithm}[htb]
	\caption{Differential Evolution (DE).}
	\label{alg:DE}
	\begin{algorithmic}[1]
		\State Set parameters $C_r, F$ and $N_p$ and iteration counter $g \leftarrow 0$
		\State Initialize population $P_{x,g} = (x_{i,g}), i = 0,1,\cdots,N-1$
		\State $fit \leftarrow$ evaluate($x_g$)
		\While {Stop Criteria is False}
		\For {i = 0 to pop\_size-1}
		\State Select a index number: $r_0$
		\State $v_{i,g} \leftarrow$ differential\_mutation($F, x_{i,g}, r_0$)
		\State $u_{i,g} \leftarrow$ recombination($Cr, v_{i,g}, x_{i,g}$)
		\State $fit'\ \leftarrow$ evaluate($u_{i,g}$)
		\If {$fit'\ < fit_i$}
		\State $x_{i,g+1} \leftarrow u_{i,g}$
		\State $fit_i \leftarrow fit'\ $
		\Else
		\State $x_{i,g+1} \leftarrow x_{i,g}$
		\EndIf
		\EndFor
		\EndWhile
	\end{algorithmic}
\end{algorithm}

In Algorithm \ref{alg:DE}, the loop between line 5 and line 14 can be calculated parallelly. Suppose we denote a parallel calculation task as a "job", meaning that the device we used can do all the jobs parallelly. Hence, each iteration of the loop between line 5 and line 16 can be allocated to a job. In addition, in line 7, line 8, and line 12, we need to loop all the elements of the chromosome vector, since each loop is independent. However, it is not good enough to set sub-jobs in a parallel calculation job: As we can't control the running time of line 10 (evaluation), it's not good to allocate each loop from line 5 to line 16 to a single job for a thread to do by including a time-consuming evaluation task in it. It's necessary to change the algorithm flow design of DE to a new one where the procedure of line 10 is separated so that other procedures of it can be calculated parallelly on MKL/CUDA. The pseudocode of the parallel DE framework can be seen in Algorithm \ref{alg:parallel_DE}
\begin{algorithm}[htb]
	\caption{Parallel Differential Evolution framework.}
	\label{alg:parallel_DE}
	\begin{algorithmic}[1]
		\State Set parameters $C_r, F$ and $N_p$ and iteration counter $g \leftarrow 0$
		\State Initialize population chromosome matrix $x_g$
		\State $fit \leftarrow$ evaluate($x_g$)
		\While {Stop Criteria is False}
		\State shuffle the order of the individuals
		\State $r_0 \leftarrow$ base\_index\_selection($N_p$)
		\State $mask \leftarrow$ crossover($Cr, N_p, D$)
		\State $u_g \leftarrow$ copy($x_g$)
		\State $u_g[mask] \leftarrow$ differential\_mutation($F, x_g, mask, r_0$)
		\State $fit'\ \leftarrow$ evaluate($u_g$)
		\State $idx \leftarrow$ where($fit'\ < fit$)
		\State $x_{g+1} \leftarrow$ copy($x_g$)
		\State $x_{g+1}[idx,:] \leftarrow u_g[idx,:]$
		\State $fit[idx] \leftarrow fit'\ [idx] $
		\EndWhile
	\end{algorithmic}
\end{algorithm}

In Algorithm \ref{alg:parallel_DE}, $r_0$ in line 6 is a vector that stores the indices of individuals chosen by the selection method. The base vectors in DE are equal to $x_g[r_0,:]$. $r_0$ can be created parallelly. In line 7, $mask$ is a 0-1 matrix storing the message whether which element of the $u_g$ matrix needs to be changed its value to the mutation result or not(1 means yes while 0 means no). It's easy to see that the order between the crossover and the mutation is exchanged, which can decrease the running time but get the same result as the traditional DE. "DE/*/*/bin" can be designed according to Algorithm \ref{alg:parallel_DE}. This framework can be used in many different versions of DE. For example, as to the "DE/*/*/bin", we can use this framework parallelly. Except line 10 (evaluation), all other codes from line 5 to line 14 can be executed parallelly by matrix calculation with the help of MKL/CUDA, which means that in many underlying operating procedures, for example, how many sub-threads are used in the parallel calculation, is setting to a suitable state by MKL/CUDA automatically. There are many kinds of base index selection in DE, some can be seen in Algorithm \ref{alg:selection1}, \ref{alg:selection2}, and \ref{alg:selection3} The basic parallel mutation of "DE/*/1/bin" can be seen in Algorithm \ref{alg:parallel_mutation}, and the parallel uniform (binomial) crossover can be seen in Algorithm \ref{alg:parallel_uniform_crossover}.

\begin{algorithm}[htb]
	\caption{Random Base Index Selection.}
	\label{alg:selection1}
	\begin{algorithmic}[1]
		\Require
		$N_p$.
		\Ensure
		An array\_like $r_0$.
		\State $r_0 \leftarrow N_p\cdot$floor(rand(0,1,size=$N_p$))
	\end{algorithmic}
\end{algorithm}

\begin{algorithm}[htb]
	\caption{Random Offset Base Index Selection.}
	\label{alg:selection2}
	\begin{algorithmic}[1]
		\Require
		$N_p$.
		\Ensure
		An array\_like $r_0$.
		\State $idx \leftarrow [0,1,\cdot,N_p-1]$
		\State $r_g \leftarrow$ ceil($(N_p-1)\cdot$rand(0,1))
		\State $r_0 \leftarrow (idx+r_g)$ mod $N_p$
	\end{algorithmic}
\end{algorithm}

\begin{algorithm}[htb]
	\caption{Permutation Base Index Selection.}
	\label{alg:selection3}
	\begin{algorithmic}[1]
		\Require
		$N_p$.
		\Ensure
		An array\_like $r_0$.
		\State $idx \leftarrow [0,1,\cdot,N_p-1]$
		\State $r_0 \leftarrow$ shuffle($idx$)
	\end{algorithmic}
\end{algorithm}

The three selections above can all be easily calculated on MKL/CUDA.

\begin{algorithm}[htb]
	\caption{Basic Parallel Differential Mutation.}
	\label{alg:parallel_mutation}
	\begin{algorithmic}[1]
		\Require
		$F, x_g, mask, r_0$.
		\Ensure
		An array\_like $results$.
		\State $N_p, Lind \leftarrow$ the size of $x_g$
		\State $r_1 \leftarrow r_0+(N_p-1)\cdot$ceil(rand(0,1,size=$N_p$)) mod $N_p$
		\State $r_2 \leftarrow r_1+(N_p-1)\cdot$ceil(rand(0,1,size=$N_p$)) mod $N_p$
		\State $results \leftarrow (x_g[r0, :])[mask] + F\cdot((x_g[r1, :])[mask] - (x_g[r2, :])[mask]) $
	\end{algorithmic}
\end{algorithm}

\begin{algorithm}[htb]
	\caption{Parallel Uniform (Binomial) Crossover (MKL/CUDA-BIN).}
	\label{alg:parallel_uniform_crossover}
	\begin{algorithmic}[1]
		\Require
		$Cr\in[0,1],N_p,D$.
		\Ensure
		$mask$.
		\State $mask \leftarrow$ an $N_p\times D$ matrix in which all elements are 0
		\State $j_{rand} \leftarrow$ randint($D$, size=($N_p$))
		\State $idx \leftarrow [0,1,\cdots,N_p-1]$
		\State $mask[idx,j_{rand}] \leftarrow 1$
		\State $R \leftarrow$ rand(0,1,size=($N_p,D$))
		\State $mask[R<Cr] \leftarrow 1$
	\end{algorithmic}
\end{algorithm}

The Altorithm \ref{alg:parallel_mutation} implements the differential mutation in the "DE/*/1/bin", the formula is as follows:
\begin{equation}
	u_i \leftarrow x_{r0} + F\cdot (x_{r1}-x_{r2})
\end{equation}

In traditional DE, the crossover is executed after mutation. After the crossover is done, quite many elements of $v_{i,g}$ which cannot be given to $u_{i,g}$ are wasted. So the $mask$ helps to speed up DE.

The Algorithm \ref{alg:parallel_uniform_crossover} implements the uniform (binomial) crossover, the formula is as follows:
\begin{equation}
	{u_{i,j,g}} \leftarrow \left\{ {\begin{array}{*{20}{c}}
			{{v_{i,j,g}},\quad if\quad rand(0,1) < Cr\quad or\quad j=j_{rand}}\\
			{{x_{i,j,g}},\quad otherwise}
	\end{array}} \right.
\end{equation}

$j$ is the index number which values in ${0,1,\cdots,D-1}$. the $mask$ is a 0-1 matrix that marks whether the elements of $u_g$ matrix which are need to be crossed with $v_g$ matrix or not.

The algorithms above can easily be done in MKL/CUDA, because most procedures of them are mainly matrix calculation. In section 4 we will show the acceleration under MKL/CUDA.

\section{New Exponential Crossover}

\subsection{The Essence of the Exponential Crossover}
The exponential crossover in Differential Evolution is similar to 1 or 2 point crossover in the Genetic Algorithm (GA) \cite{Storn1995}, which is widely used in "DE/*/*/L". Its implementation is shown in Algorithm \ref{alg:exponential_crossover}. In each generation $g$, a mutation vector $v_{i,g}$ created from the parent chromosome vector $x_{i,g}$ by applying some mutation strategies. After that, the mutation vector $v_{i,g}$ is crossed with $x_{i,g}$ in order to create trial vector $u_{i,g}$ by applying the exponential crossover. $Cr \in \left[ {0,1} \right]$ is the crossover rate, $j_{rand}$ is a decision variable index randomly selected from [0,$D-1$].

\begin{algorithm}[htb]
	\caption{Exponential Crossover.}
	\label{alg:exponential_crossover}
	\begin{algorithmic}[1]
		\Require $x_g,v_g,Cr\in[0,1]$
		\Ensure $u_g$
		\State $N_p,D \leftarrow$ the shape of $x_g$
		\For {$i \leftarrow [0,1,\cdots,N_p-1]$}
		\State $u_{i,g} \leftarrow x_{i,g}$
		\State $j$ is randomly selected from [0,$D$-1], $L=0$
		\Do
		\State $u_{i,j,g} \leftarrow v_{i,j,g}$
		\State $j \leftarrow (j+1)$ mod $D$
		\State $L \leftarrow L+1$
		\DoWhile {rand[0,1)<$Cr$ and $L<D$}
		\EndFor
	\end{algorithmic}
\end{algorithm}

\begin{figure}[h]
	\centering
	\includegraphics[scale=0.5]{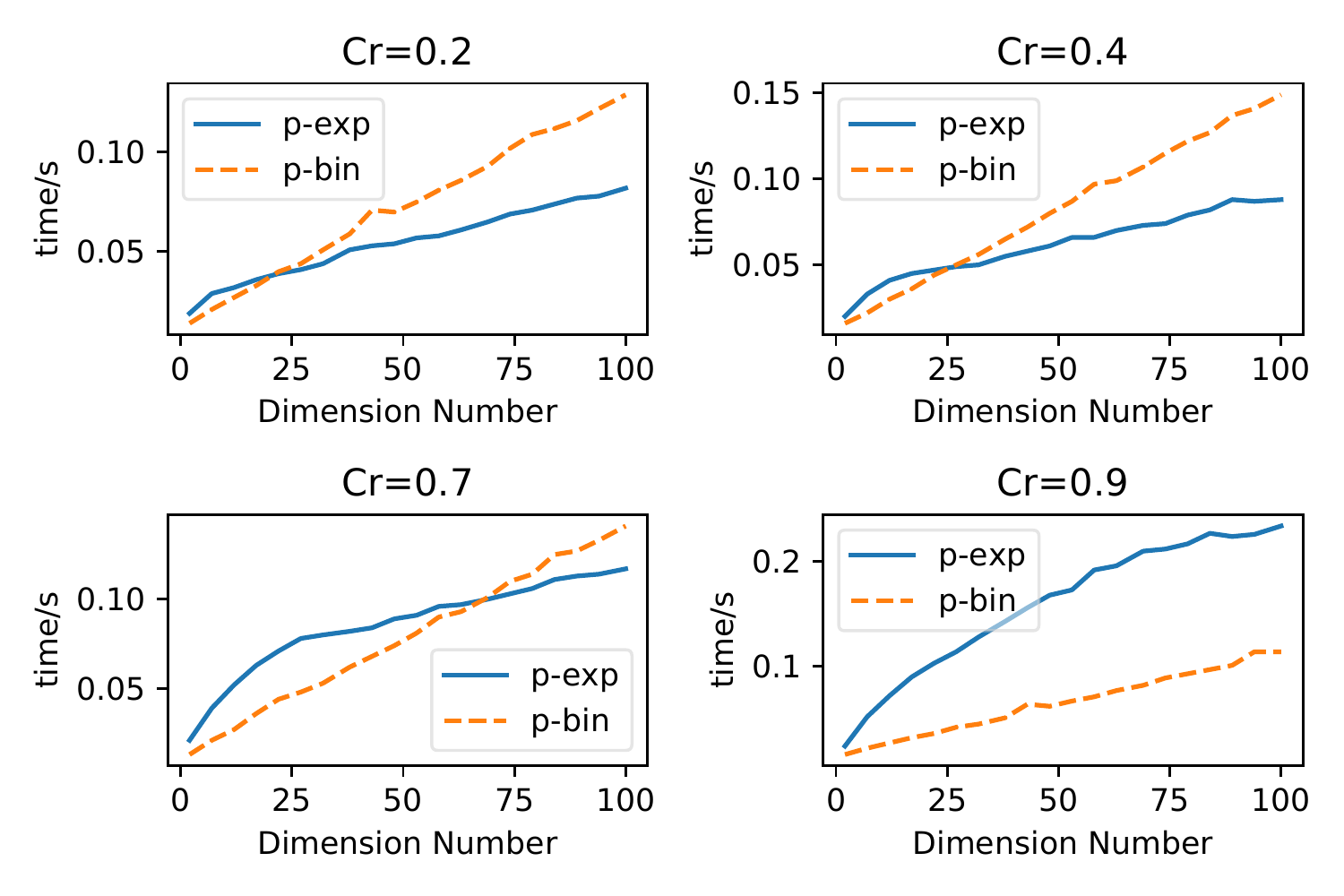}
	\caption{Running Time Comparison between p-exp and p-bin in $N_p=1000$}
	\label{fig:Comparison1}
\end{figure}

However, the exponential crossover is seldom used in parallel DE practices. We can easily find the reason. It's clear that if we allocate each execution of the exponential crossover of each individual as a parallel job, that each job needs to loop the chromosome vector to check whether $u_{i,g}$ is needed to be crossed. This procedure cannot run parallelly. As to the uniform (binomial) crossover, the process of checking whether bit of chromosome is needed to be crossed or not is independent. Thus, the latter is more suitable for parallel calculation. Figure \ref{fig:Comparison1} shows the runtime comparison between parallel exponential crossover (p-exp) and parallel uniform (binomial) crossover (p-bin).

The exponential crossover has a congenital advantage that when $Cr$ is relatively small and the dimension of $u_{i,g}$ is large, it doesn't need to traverse all bit of $u_{i,g}$ to cross, because the loop will be probability stopped on the half-way. So we can see in Figure \ref{fig:Comparison1} that when $Cr=0.2$ and $Cr=0.4$, as the dimension of the chromosome vector is getting larger, the p-exp has a higher performance than p-bin. However, when $Cr$ is getting larger, p-exp is slower than p-bin. That is because each parallel job of p-exp has more work to do: it needs to traverse the bits of $u_{i,g}$ to check whether it is needed to cross.

In some related work, the $L$ of each $v_{i,g}$ can be calculated before the crossover, so that remain procedures of the exp crossover can be calculated parallelly at a higher speed. That can be seen in \cite{Zhao2013}. However, when calculating the $L$, they still use a loop same to the traditional exp crossover, so that these improvements are still unsuitable for parallel calculation.

If we reconsider the exponential crossover, we can see that the $L$, which is equal to the length of crossed bits in $u_{i,g}$, is subject to geometric distribution, not the exponential distribution. But as the geometric distribution has some relationship with the exponential distribution, it's unnecessary to change the name of the exponential crossover. Instead, we can use the probability theory to redesign the exponential crossover in order to make it faster.

According to Algorithm \ref{alg:exponential_crossover}, we can get
\[L \sim G\left( {1 - Cr} \right)\ \]

Where $G$ denotes the geometric distribution, meaning that
\begin{equation}
	P\left\{ {L = n} \right\} = C{r^{n - 1}}\left( {1 - Cr} \right),n \ge 1
\end{equation}

What we use here is the property that
\begin{equation}
	G\left( n \right) = P\left\{ {L \le n} \right\} = 1 - \sum\limits_{i > n} {\left( {1 - Cr} \right)C{r^{i - 1}} = 1 - C{r^n}}
\end{equation}

Now we need to generate a random number that is subject to the geometric distribution. Denote X as an exponentially distributed random variable with parameter $\lambda$, so we have $L = \left\lceil X \right\rceil$, where $\left\lceil\quad\right\rceil$ is the ceil (or smallest integer) function. $L$ is a geometrically distributed random variable with parameter $p=1-e^{-\lambda}$. $p$ is the parameter of the geometric distribution. In the exponential crossover, $p$ is equal to $1-Cr$. Thus, we obtain $\lambda  =  - \ln Cr$. If we denote $F_X(x)$ as the distribution function of the exponential distribution, we can get
\begin{equation}
	\label{equ:exp}
	F_X(x)=1-Cr^x
\end{equation}

According to the inversion method \cite{Devroye2010} and equation \ref{equ:exp}, we can get
\begin{equation}
	U_1=F_X(x)
\end{equation}

where $U_1$ is a random number in (0,1). Is clear that $U=1-U_1$ is also a uniform distribution random number in (0,1). Thus, we can get a random sampling real number $x$ which is exponentially distributed.
\begin{equation}
	x = \frac{{\ln \left( U \right)}}{{\ln \left( {Cr} \right)}}
\end{equation}

So, the $L$ in the exponential crossover can be sampled randomly by the following equation instead of traversing elements of the chromosome vector.

\begin{equation}
	L = \left\lceil {\frac{{\ln U}}{{\ln Cr}}} \right\rceil
\end{equation}

\subsection{New Exponential Crossover (NEC)}
In the new exponential crossover, since $Cr\in[0,1]$, it is necessary to duel with the three special cases: $Cr=0$, $Cr=1$, and $L$ cannot be bigger than the dimension number $D$ of the $u_{i,g}$. Thus, the equation to generate $L$ is as follows.

\begin{equation}
	\label{equ:n-exp}
	L = \left\{ {\begin{array}{*{20}{l}}
			{1,Cr = 0}\\
			{\min \left( {\left\lceil {\frac{{\ln U}}{{\ln Cr}}} \right\rceil ,D} \right),Cr \in \left( {0,1} \right)}\\
			{D,Cr = 1}
	\end{array}} \right.
\end{equation}

After generating $L$, we can know which bits of $u_{i,g}$ are needed to be crossed directly. So we can redesign the exponential crossover. Algorithm \ref{alg:new_exponential_crossover} shows the new design of the exponential crossover.
\begin{algorithm}[htb]
	\caption{New Exponential Crossover (NEC).}
	\label{alg:new_exponential_crossover}
	\begin{algorithmic}[1]
		\Require $x_g,v_g,Cr\in[0,1]$
		\Ensure $u_g$
		\State $N_p,D \leftarrow$ the shape of $x_g$
		\For {$i \leftarrow [0,1,\cdots,N_p-1]$}
		\State $u_{i,g} \leftarrow x_{i,g}$
		\State $j$ is randomly selected from [0,$D$-1], $l=0$
		\State calculate $L$ by equation \ref{equ:n-exp}
		\Do
		\State $u_{i,j,g} \leftarrow v_{i,j,g}$
		\State $j \leftarrow (j+1)$ mod $D$
		\State $l \leftarrow l+1$
		\DoWhile {$l<L$}
		\EndFor
	\end{algorithmic}
\end{algorithm}

\begin{figure}[h]
	\centering
	\includegraphics[scale=0.5]{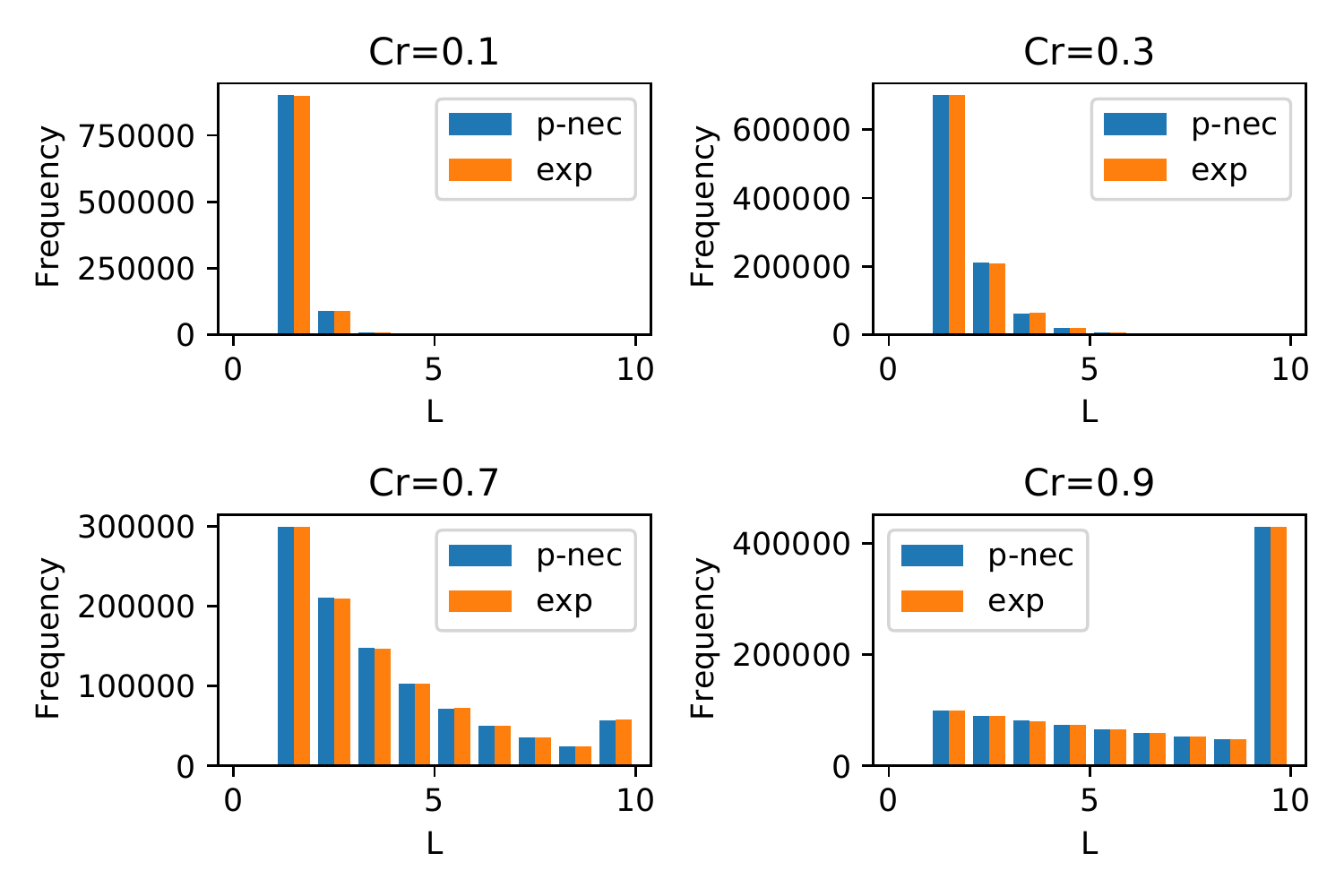}
	\caption{The frequencies of $L$ between nec and exp}
	\label{fig:Comparison2}
\end{figure}

Figure \ref{fig:Comparison2} shows the frequency of $L$ in the new exponential crossover (NEC) and the traditional exponential crossover (exp) in 1000000 times experiments with $D=10$. We can see that we can get similar results as the traditional exponential crossover by using NEC.

\begin{figure}[h]
	\centering
	\includegraphics[scale=0.5]{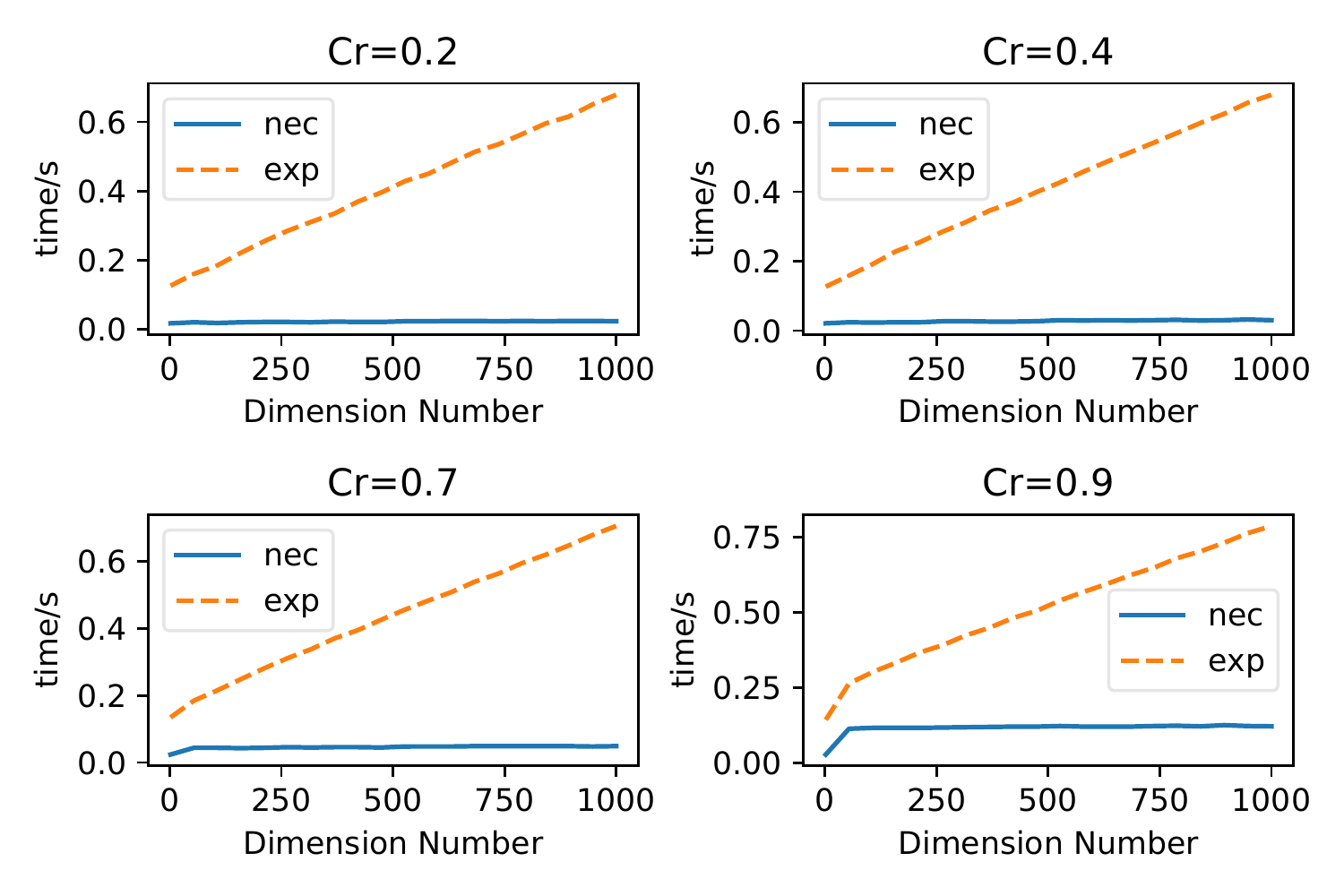}
	\caption{Running Time Comparison between nec and exp in $N_p=1000$}
	\label{fig:Comparison3}
\end{figure}

\begin{figure}[h]
	\centering
	\includegraphics[scale=0.5]{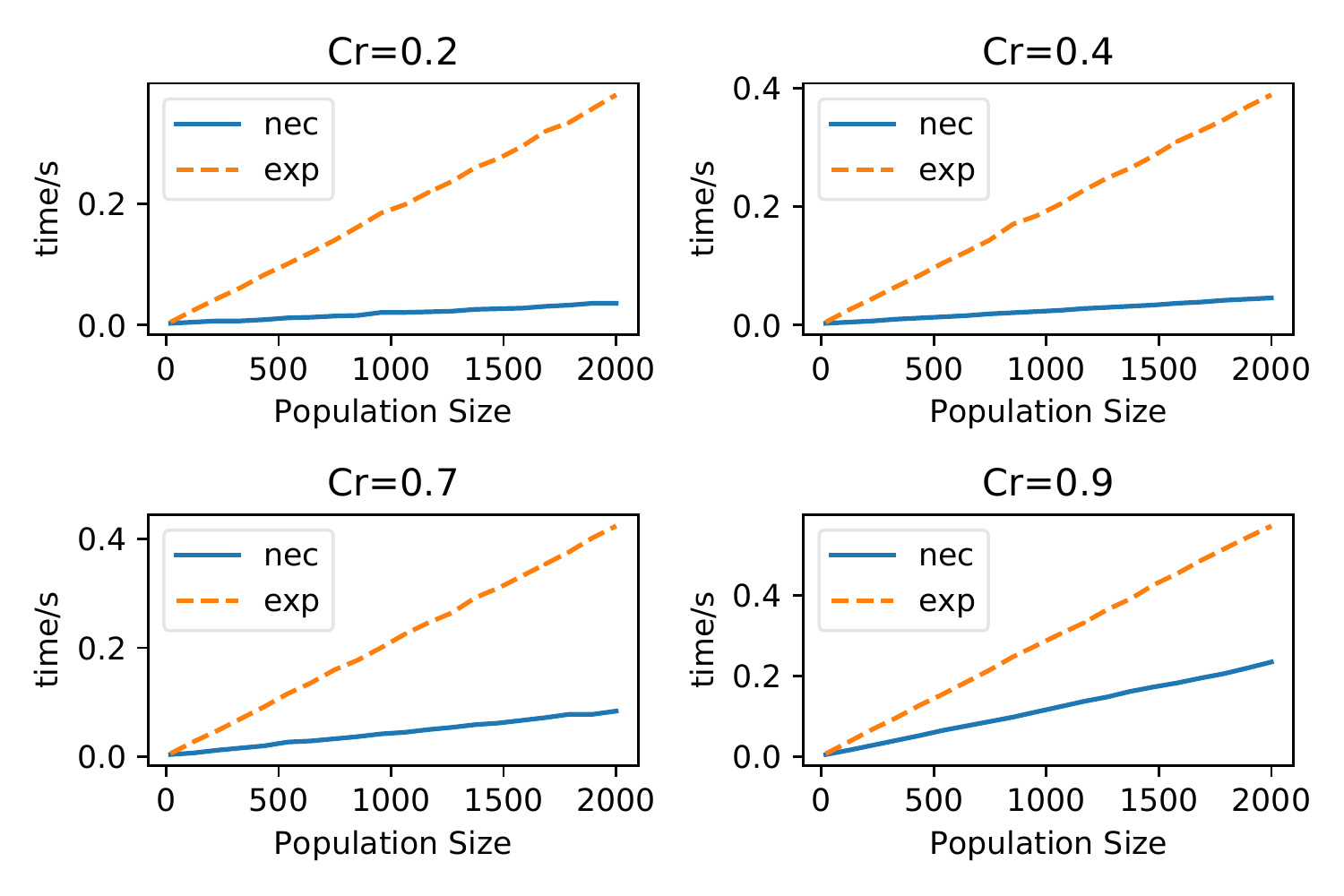}
	\caption{Running Time Comparison between nec and exp in $D=100$}
	\label{fig:Comparison3_1}
\end{figure}

Figure \ref{fig:Comparison3} and \ref{fig:Comparison3_1} show the higher speed of NEC in 100 experiments with different $Cr$.

Comparing the traditional exponential crossover and NEC, the latter calculates the total crossed length $L$ at first so that when looping $u_{i,g}$, it only needs to compare the crossed length $l$ with $L$, which makes it much faster than the traditional exponential crossover.

Moreover, the procedure of generating all $L$ of all individuals in DE can be also executed parallelly, so that we can make full use of MKL/CUDA to speed up the procedure of the new exponential crossover. Firstly, the column vector $Ls$, that stores all the $L$ of the $u_{i,g}, i=0,1,2,\cdots,N_p$ are sampled parallelly by applying equation \ref{equ:n-exp}. Then, with the help of the $Ls$, we can create a matrix $mask$ parallelly which denotes whether every bit of all the $u_{i,g}$ of all chromosome vectors need to be crossed or not. Finally, we can do the cross of every bit of the chromosomes parallelly on MKL/CUDA according to the matrix $mask$. It is worth noting that the new exponential crossover on MKL/CUDA (MKL/CUDA-NEC) only returns a signal matrix $mask$, which is similar to Algorithm \ref{alg:parallel_uniform_crossover}, that is because in parallel differential evolution on MKL/CUDA (see Algorithm \ref{alg:parallel_DE}), the mutation is executed after the crossover. With the help of the $mask$, the mutation only needs to mutate the elements where the element of the $mark$ is equal to 1. Algorithm \ref{alg:parallel_new_exponential_crossover} is the pseudocode of the parallel new exponential crossover (p-nec).

\begin{algorithm}[htb]
	\caption{Parallel Version of New Exponential Crossover (MKL/CUDA-NEC).}
	\label{alg:parallel_new_exponential_crossover}
	\begin{algorithmic}[1]
		\Require
		$Cr\in[0,1],N_p,D$.
		\Ensure
		$mask$.
		\State $j_{rand} \leftarrow$ randint($D$, size=($N_p$))
		\State create a column vector $Ls$ by applying equation \ref{equ:n-exp} parallelly
		\State $Seq \leftarrow$ repmat($[0,1,\cdots,D-1]$, size=($N_p$,1))
		\State $j \leftarrow$ repmat($j_{rand}$,size=(1,$D$))
		\State $j'\ \leftarrow$ repmat($Ls$,size=(1,$D$))
		\State $mask \leftarrow (j<=Seq)$ XOR $(j'\ <Seq)$
	\end{algorithmic}
\end{algorithm}

\begin{figure}[h]
	\centering
	\includegraphics[scale=0.5]{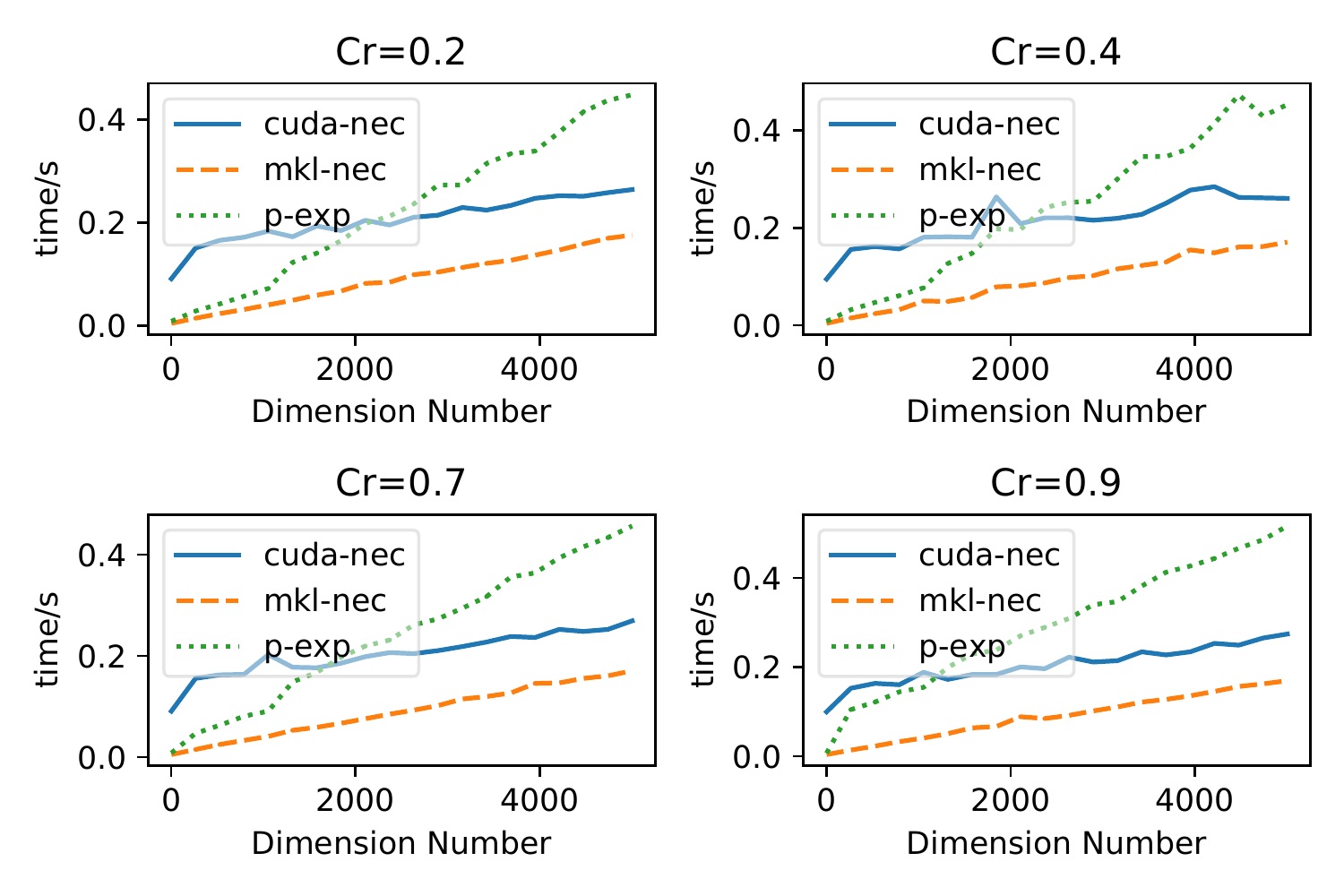}
	\caption{Running Time Comparison among CUDA-nec, MKL-nec, and p-exp when $N_p=100$}
	\label{fig:Comparison4}
\end{figure}

\begin{figure}[h]
	\centering
	\includegraphics[scale=0.5]{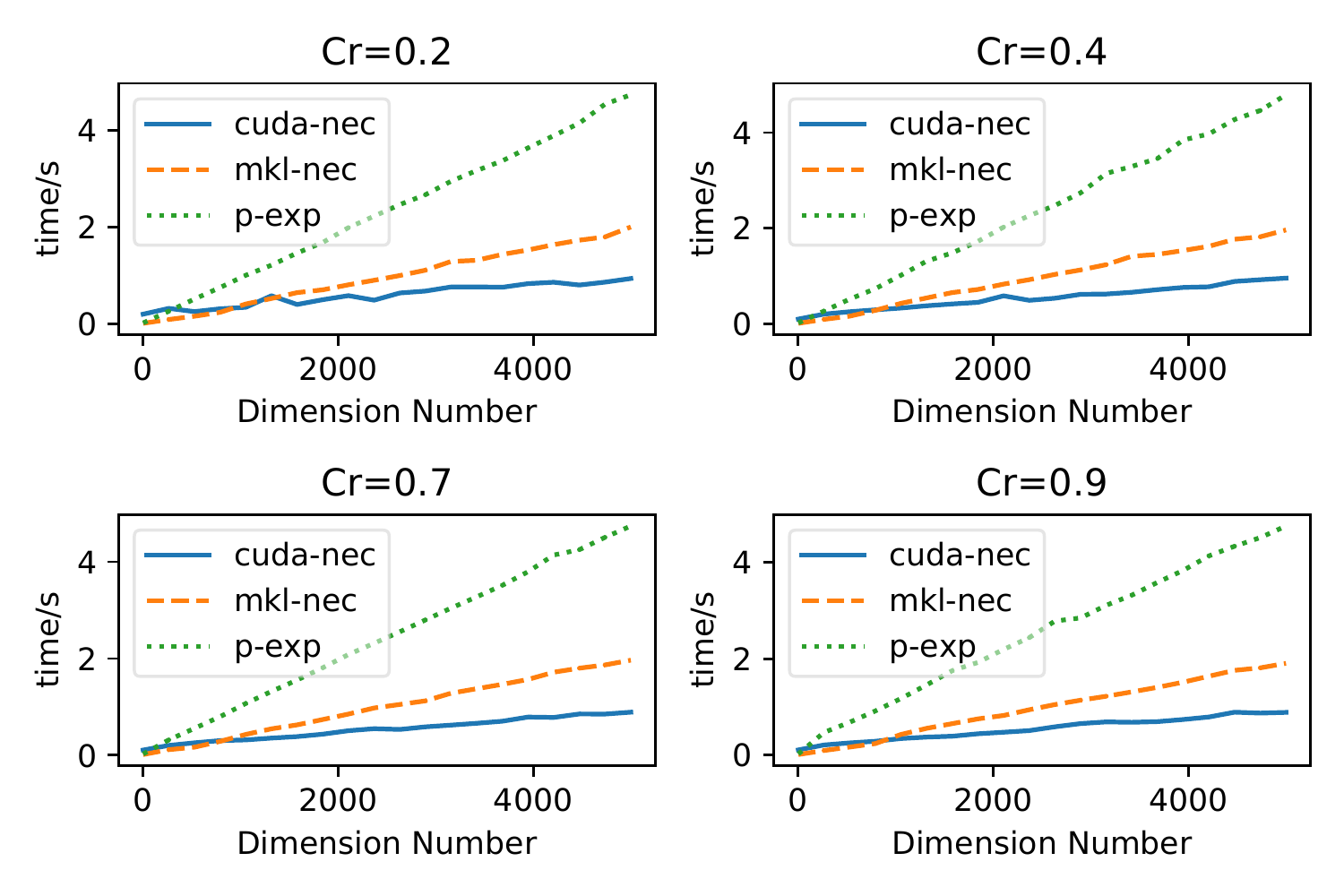}
	\caption{Running Time Comparison among CUDA-nec, MKL-nec, and p-exp when $N_p=1000$}
	\label{fig:Comparison5}
\end{figure}

\begin{figure}[h]
	\centering
	\includegraphics[scale=0.5]{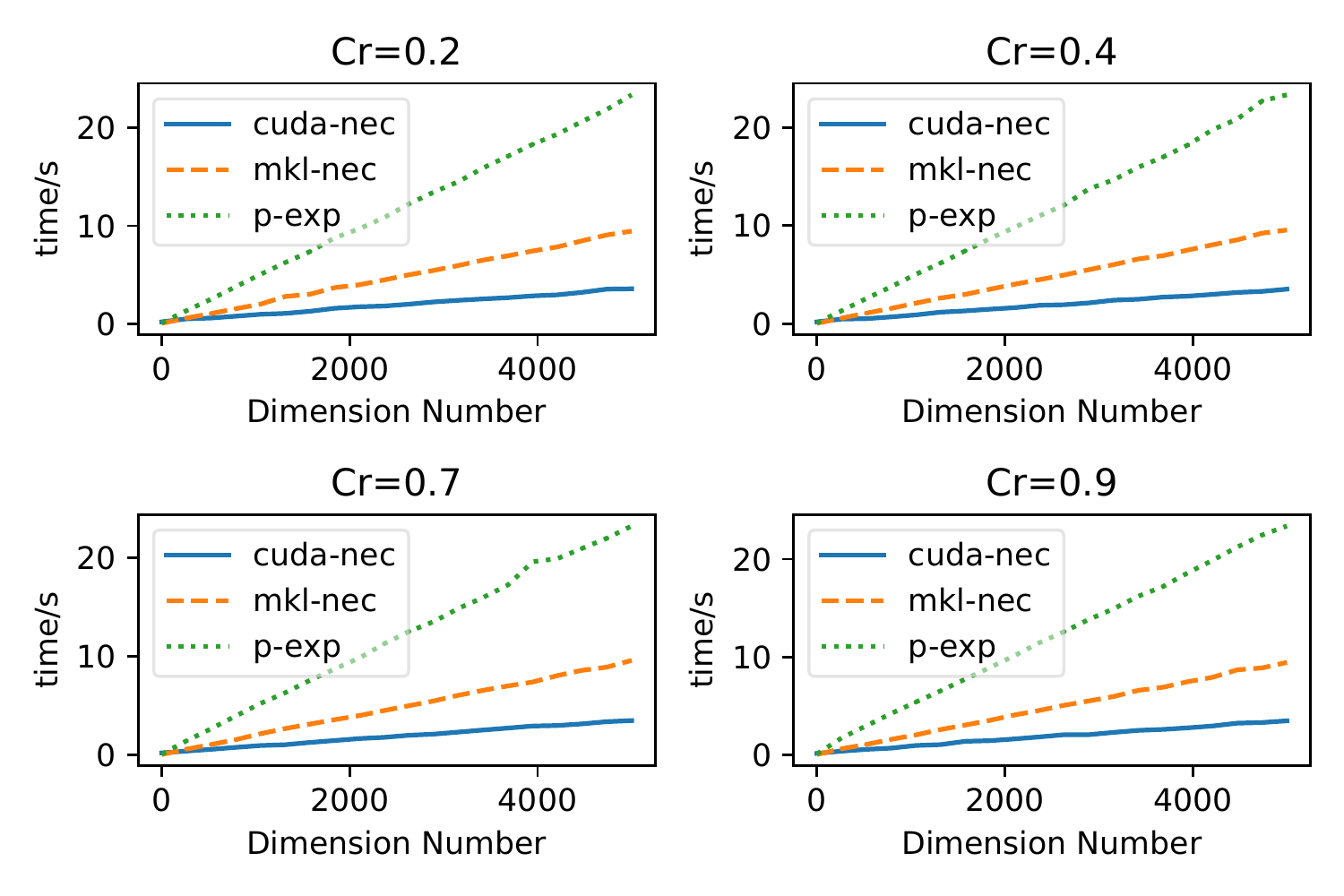}
	\caption{Running Time Comparison among CUDA-nec, MKL-nec, and p-exp when $N_p=5000$}
	\label{fig:Comparison6}
\end{figure}

Comparing MKL/CUDA-nec to p-exp we mentioned above, we can see that all the procedures of the former one are the matrix calculation, which can be easily run thoroughly parallelly with the help of MKL/CUDA. Thus, it runs much faster than p-exp. Figure \ref{fig:Comparison4}, Figure \ref{fig:Comparison5}, and Figure \ref{fig:Comparison6} will show the runtime comparison among CUDA-nec (using CUDA to calculate NEC), MKL-nec (using MKL to calculate NEC), and p-exp when the population size $N_p$ is 100, 1000 and 5000 in 100 repeat experiments. The $D$ (dimension of $u_{i,g}$) is set from 2 to 5000. We can see from these pictures that when $N_p$ is small, MKL-NEC is faster than the others, but when $N_p$ is large and the dimension number of the chromosome is getting larger and larger, CUDA-NEC can execute faster. That is because in big optimization, the runtime of data's transport between the memory device and the GPU is no longer the bottleneck, thus, we get a higher speedup by using GPU to calculate.

Table \ref{tab:comparison-time} shows more experiment results. Each experiment are done in $Cr$=0.2, 0.4, 0.6, 0.8 and 1.0 for 100 times then records the total runtime. We can see that when the dimension of the decision variable ($D$) and the population size ($N_p$) is relatively small, the new exponential crossover on MKL has a higher performance. But when $D$ and $N_p$ are relatively large, it's better to use CUDA-NEC as well as CUDA-bin. Moreover, no matter MKL is used or not, when $D$ and $N_p$ are getting larger, the uniform (binomial) crossover costs more time than the new exponential crossover. That is because, in the uniform (binomial) crossover, we need to create an $N_p\times D$ matrix in which elements are random number, while as to the new exponential crossover, with the help of equation \ref{equ:n-exp}, we only need to create a random column vector which length is $N_p$. But if we use CUDA to speed up by GPU when the computation scale is large, the runtime gap between CUDA-nec and CUDA-bin is not so significant.

\begin{table}[h]
	\centering
	\caption{Running times' comparison of several DE Crossovers}
	\setlength{\tabcolsep}{0.6mm}
	\begin{tabular}{llccccccc}
		\hline
		D & Np & exp & nec & bin & \makecell[tl]{MKL\\-nec} & \makecell[tl]{CUDA\\-nec} & \makecell[tl]{MKL\\-bin} & \makecell[tl]{CUDA\\-bin} \\ \hline
		10 & 100 & 0.05 & 0.03 & 0.06 & \cellcolor{gray95}0.02 & 1.76 & \cellcolor{gray95}0.02 & 6.99 \\ 
		10 & 1000 & 0.44 & 0.12 & 0.51 & \cellcolor{gray95}0.05 & 1.18 & \cellcolor{gray25}0.07 & 0.64 \\ 
		10 & 10000 & 4.34 & 1.17 & 4.99 & \cellcolor{gray95}0.31 & 1.34 & \cellcolor{gray25}0.53 & 0.94 \\ 
		100 & 100 & 0.11 & 0.08 & 0.23 & \cellcolor{gray95}0.03 & 1.16 & \cellcolor{gray25}0.05 & 0.66 \\ 
		100 & 1000 & 1.09 & 0.61 & 2.18 & \cellcolor{gray95}0.12 & 1.23 & \cellcolor{gray25}0.32 & 0.64 \\ 
		100 & 10000 & 10.86 & 6.15 & 22.57 & \cellcolor{gray95}0.98 & \cellcolor{gray25}1.11 & 4.18 & 1.36 \\ 
		1000 & 100 & 0.75 & 0.57 & 1.9 & \cellcolor{gray95}0.13 & 0.66 & \cellcolor{gray25}0.32 & 0.56 \\ 
		1000 & 1000 & 7.53 & 5.44 & 19.16 & \cellcolor{gray95}0.73 & 1.07 & 3.98 & \cellcolor{gray25}1.00 \\ 
		1000 & 10000 & 74.77 & 54.52 & 192.46 & 9.16 & \cellcolor{gray95}4.31 & 41.76 & \cellcolor{gray25}4.56 \\ 
		10000 & 100 & 7.1 & 5.46 & 18.67 & \cellcolor{gray25}0.99 & 1.48 & 4.1 & \cellcolor{gray95}0.97 \\ 
		10000 & 1000 & 70.82 & 53.58 & 187.6 & 7.97 & \cellcolor{gray25}4.43 & 41.49 & \cellcolor{gray95}3.91 \\ 
		10000 & 10000 & 710.4 & 538.05 & 1868.47 & 77.82 & \cellcolor{gray95}33.26 & 421.17 & \cellcolor{gray25}39.14 \\ \hline
	\end{tabular}
	\label{tab:comparison-time}
\end{table}

\section{Performance Evaluation of DE based on MKL/CUDA}

Our DE programs are implemented in Python. The MKL is implemented by Numpy-MKL and the GPU environment is set up in Cupy \cite{Okuta2017} with CUDA 10.0. In order to overcome the performance deficiency of Python, the key codes are implemented in Cython, which is the C-Extensions for Python. There are mainly three different versions of DE in this experiment: no-parallel DE, DE on MKL (MKL-DE), and DE in CUDA (CUDA-DE). In addition, as to the DE, we arrange two sub-versions: "DE/rand/1/bin" and "DE/rand/1/L". It is worth to note that the crossover of "DE/rand/1/L" is the parallel version of the new exponential crossover mentioned in Algorithm \ref{alg:parallel_new_exponential_crossover}. All of them are running on an Intel\textsuperscript{\textregistered} i5-9600K, 16GB memory, and a common NVIDIA GeForce$^\text{TM}$ GTX 760 GPU.

In this experiment, we focus our attention on three well-known benchmark functions: Ackley, Griewank and Rosenbrock \cite{Price2005}:

\begin{itemize}
	\item Ackley function: multimodal, separable,
	\[f\left( x \right) =  - 20{e^{ - 0.02\sqrt {\frac{1}{D}\sum\limits_{j = 1}^D {x_j^2} } }} - {e^{\frac{1}{D}\sum\limits_{j = 1}^D {\cos 2\pi {x_j}} }} + 20 + e\]
	${x_j} \in \left[ { - 30,30} \right],{x^*} = \left( {0,0, \cdots ,0} \right),f\left( {{x^*}} \right) = 0.$
	\item Griewank function: multimodal, nonseparable,
	\[f\left( x \right) = \sum\limits_{j = 1}^D {\frac{{x_j^2}}{{4000}}}  - \prod\limits_{j = 1}^D {\cos \left( {\frac{{{x_j}}}{{\sqrt j }}} \right)}  + 1\]
	${x_j} \in \left[ { - 400,400} \right],{x^*} = \left( {0,0, \cdots ,0} \right),f\left( {{x^*}} \right) = 0.$
	\item Rosenbrock function: multimodal, separable,
	\[f\left( x \right) = \sum\limits_{j = 1}^{D - 1} {\left[ {100{{\left( {x_j^2 - {x_{j + 1}}} \right)}^2} + {{\left( {1 - {x_j}} \right)}^2}} \right]} \]
	${x_j} \in \left[ { - 5.12,5.12} \right],{x^*} = \left( {1,1, \cdots ,1} \right),f\left( {{x^*}} \right) = 0.$
\end{itemize}

The parameters setting of DEs in the experiment can be seen in Table \ref{tab:parameters-setting}, where the $F$ and $Cr$ are referred to K. Price and R. Storn's work \cite{Price2005}. The dimension $D$ of these three benchmark functions and the population size $N_p$ are set to different values in order to compare the calculational performance in different computational scales. The results can be seen in the tables range from Table \ref{tab:Ackley-result} to Table \ref{tab:Rosenbrock-result}.

\begin{table}[h]
	\centering
	\caption{Parameters Setting}
	\setlength{\tabcolsep}{1mm}
	\begin{tabular}{lccccc}
		\hline
		Function & D & Np & F & Cr & MAX GEN \\ \hline
		Ackley & 1000 & 500 & 0.5 & 0.2 & 30000 \\
		Griewank & 500 & 300 & 0.25 & 0.45 & 20000 \\
		Rosenbrock & 50 & 100 & 0.65 & 0.95 & 10000 \\ \hline
	\end{tabular}
	\label{tab:parameters-setting}
\end{table}

\begin{table}[h]
	\centering
	\caption{Comparison of variant DEs in 1000-D Ackley with 500 individuals and 30000 generations}
	\setlength{\tabcolsep}{0.3mm}
	\begin{tabular}{lcccc}
		\hline
		Algorithms & best value & best gen & time/s & speedup \\ \hline
		DE/rand/1/L & 2.32E-05 & 30000 & 2742.93($\pm$105) & - \\
		MKL-DE/rand/1/L & 2.09E-05 & 29997 & 2314.79($\pm$52) & 1.18 \\
		CUDA-DE/rand/1/L & 2.62E-05 & 29991 & 323.97($\pm$5) & 8.47 \\
		rand/1/bin & 2.15E-04 & 29998 & 3323.53($\pm$93) & - \\
		MKL-DE/rand/1/bin & 2.84E-04 & 29951 & 2784.57($\pm$98) & 1.19 \\
		CUDA-DE/rand/1/bin & 2.28E-04 & 29983 & 321.45($\pm$11) & 10.34 \\ \hline
	\end{tabular}
	\label{tab:Ackley-result}
\end{table}

\begin{table}[h]
	\centering
	\caption{Comparison of variant DEs in 500-D Griewank with 300 individuals and 20000 generations}
	\setlength{\tabcolsep}{0.3mm}
	\begin{tabular}{lcccc}
		\hline
		Algorithms & best value & best gen & time/s & speedup \\ \hline
		DE/rand/1/L & 8.88E-16 & 18366 & 563.84($\pm$84) & - \\
		MKL-DE/rand/1/L & 3.71E-16 & 19855 & 475.81($\pm$45) & 1.19 \\
		CUDA-DE/rand/1/L & 1.73E-16 & 19936 & 149.03($\pm$5) & 3.78 \\
		rand/1/bin & 2.22E-16 & 8125 & 732.84($\pm$114) & - \\
		MKL-DE/rand/1/bin & 2.22E-16 & 6291 & 524.01($\pm$95) & 1.40 \\
		CUDA-DE/rand/1/bin & 2.22E-16 & 6041 & 143.95($\pm$8) & 5.09 \\ \hline
	\end{tabular}
	\label{tab:Griewank-result}
\end{table}

\begin{table}[h]
	\centering
	\caption{Comparison of variant DEs in 50-D Rosenbrock with 100 individuals and 10000 generations}
	\setlength{\tabcolsep}{0.3mm}
	\begin{tabular}{lcccc}
		\hline
		Algorithms & bestValue & best\_gen & time/s & speedup \\ \hline
		DE/rand/1/L & 2.04E-07 & 9991 & 15.43($\pm$2) & - \\
		MKL-DE/rand/1/L & 4.28E-07 & 9989 & 9.96($\pm$1) & 1.55 \\
		CUDA-DE/rand/1/L & 1.16E-07 & 9996 & 58.92($\pm$3) & 0.26 \\
		rand/1/bin & 3.71E-10 & 10000 & 19.13($\pm$2) & - \\
		MKL-DE/rand/1/bin & 2.17E-10 & 9998 & 10.01($\pm$1) & 1.91 \\
		CUDA-DE/rand/1/bin & 9.93E-10 & 10000 & 56.72($\pm$4) & 0.34 \\ \hline
	\end{tabular}
	\label{tab:Rosenbrock-result}
\end{table}

From the results, we can see that the DE (see Algorithm \ref{alg:parallel_DE}) on MKL/CUDA is much faster than traditional DE (see Algorithm \ref{alg:DE}). And between the DE on MKL (MKL-DE) and the DE in CUDA (CUDA-DE), when the dimension of the variable is relatively small, for example, $D=50$, MKL-DE is the fastest. While $D$ is relatively larger, for example, $D=500$ or $D=1000$, and the population size $N_p$ is large, CUDA-DE is fastest. That is because Algorithm \ref{alg:parallel_DE} is mainly matrix calculations, which can be executed very fast on MKL/CUDA. And when the matrix scale is quite large, the CUDA can do the calculation much faster by using GPU to accelerate.

\section{Conclusions and Future Work}
This paper firstly presents an available parallel differential evolution framework which is suitable to be built on MKL/CUDA. The procedure of this parallel DE is mainly matrix calculation so that with the help of the MKL/CUDA, the DE program can be run in a high speed even parallelly if possible. Moreover, it doesn't need to worry about the parameters of the kernel processes of the parallel calculation such as the number of CPU cores, GPU cores, threads, and so forth, these parameters are set automatically and suitably in MKL/CUDA.

Later, we analyzed the disadvantage of the exponential crossover of DE, which is inefficient and isn't suitable for MKL/CUDA to calculate totally parallelly. Hence, a new exponential crossover is presented in Algorithm \ref{alg:new_exponential_crossover}, which can run faster. Moreover, a parallel version of the new exponential crossover is proposed in Algorithm \ref{alg:parallel_new_exponential_crossover}, which is all the matrix calculation so that it can be executed in a high speed with the help of MKL/CUDA. From the experimental results presented in Section 3, we can see that the computational performance of the parallel version of the new exponential crossover catches up with the parallel binomial (uniform) crossover on MKL/CUDA.

The experimental results presented in Section 4 shows the higher performance of the parallel DE on MKL/CUDA. Especially when the dimension of the decision variable of the optimization problem is quite large, the DE on CUDA has a higher acceleration ratio.

Since differential evolution is a stochastic optimization evolution algorithm that is inherently parallel and it is similar to many other evolutionary algorithms, the whole area of evolutionary computation may also get benefit from the MKL/CUDA. In the future, we will make more effort to speed up more evolutionary algorithms.

\enlargethispage{-11.50cm}


%



\ifCLASSOPTIONcaptionsoff
  \newpage
\fi



%

\bibliographystyle{IEEETran}
\bibliography{IEEEabrv,car-ref}



%




\end{document}